\documentclass[letterpaper, 10 pt, conference]{ieeeconf}  

\let\labelindent\relax

\usepackage{enumitem}
\usepackage{graphicx}
\usepackage{color}
\usepackage{array}
\usepackage{algorithm}
\usepackage{algpseudocode}
\usepackage{amssymb}
\usepackage{soul}
\usepackage[T1]{fontenc} 
\usepackage{amsmath}
\usepackage{hyperref}

\renewcommand{\baselinestretch}{0.975}

\newlength\savedwidth
\newcommand{\wcline}[1]{\noalign{\global\savedwidth\arrayrulewidth\global\arrayrulewidth 0.75pt} \cline{#1}
\noalign{\global\arrayrulewidth\savedwidth}}

\makeatletter
 \def\Hline{%
 \noalign{\ifnum0=`}\fi\hrule \@height 1.5pt \futurelet
 \reserved@a\@xhline}
 \makeatother

\IEEEoverridecommandlockouts                              

\title{\LARGE \bf
GONet: A Semi-Supervised Deep Learning Approach\\
For Traversability Estimation}
\author{Noriaki Hirose$^{1*}$, Amir Sadeghian$^{1*}$, Marynel V\'azquez$^{1}$, Patrick Goebel$^{1}$,  and Silvio Savarese$^{1}$
\thanks{* indicates equal contribution}
\thanks{$^{1}$N. Hirose, A. Sadeghian, M. V\'azquez, P. Goebel, and S. Savarese are with the Stanford AI Lab, Computer Science Department,
        Stanford University, 353 Serra Mall, Stanford, CA, USA
        {\tt\small hirose@cs.stanford.edu}}%
}

\begin{document}
\maketitle
\thispagestyle{empty}
\pagestyle{empty}

\begin{abstract}
We present semi-supervised deep learning approaches for traversability estimation from fisheye images. Our method, GONet, and the proposed extensions leverage Generative Adversarial Networks (GANs) to effectively predict whether the area seen in the input image(s) is safe for a robot to traverse. These methods are trained with many positive images of traversable places, but just a small set of negative images depicting blocked and unsafe areas. This makes the proposed methods practical. Positive examples can be collected easily by simply operating a robot through traversable spaces, while obtaining negative examples is time consuming, costly, and potentially dangerous. Through extensive experiments and several demonstrations, we show that the proposed traversability estimation approaches are robust and can generalize to unseen scenarios. Further, we demonstrate that our methods are memory efficient and fast, allowing for real-time operation on a mobile robot with single or stereo fisheye cameras. As part of our contributions, we open-source two new datasets for traversability estimation. These datasets are composed of approximately 24h of videos from more than 25 indoor environments. Our methods outperform baseline approaches for traversability estimation on these new datasets.
\end{abstract}
%
\section{INTRODUCTION}
%

Effective identification of non-traversable spaces is essential for long-term mobile robot operation. Autonomous service robots \cite{thrun99:inproceedings, mutlu08:inproceedings, hirose13:inproceedings, veloso2015cobots}, electric wheelchairs \cite{oh08:article, hirose2010:inproceedings}, rolling walkers \cite{kulyukin08-inproceedings}, and other smart mobile machines need to move safely in dynamic environments; otherwise they could damage themselves or even injure people \cite{security-robot16:misc}. 

Common approaches to recognize non-traversable paths during robot navigation use bumpers or depth measurements \cite{papadakis13:article}. Unfortunately, bumpers don't prevent robots from falling off edges and can fail to detect small obstacles. Depth measurements can be obtained from 2D/3D lidars \cite{cinietal02:inproceedings,pfeiffer2017:inproceedings, suger15:inproceedings}, depth cameras \cite{flacco12:inproceedings, bogoslavskyi13:inproceedings}, stereo camera pairs \cite{koyasu01:inproceedings}, or single cameras with 3D reconstruction methods 
\cite{daftry16:article, alvarez2016:inbook}. But lidar sensors are expensive and depth measurements can be affected by surface textures and materials. For example, lidars and depth cameras often have trouble sensing highly reflective surfaces and transparent objects, such as mirrors and glass doors. These problems have motivated alternative approaches for traversability estimation using RGB cameras. For instance, prior work has used images to classify terrain into traversable and non-traversable areas \cite{kim07:inproceedings, ulrich00:inproceedings}. 

We contribute semi-supervised computer vision methods for traversability estimation. These methods learn to distinguish traversable and non-traversable areas by using many positive examples of places that are safe to navigate through, but just a few negative images, e.g., displaying stairs in front of a robot or objects that a robot will imminently collide with. Being able to learn from such uneven data makes our approach practical because collecting positive examples is easy -- we can drive a robot through safe areas \cite{ulrich00:inproceedings, suger15:inproceedings}. But collecting negative examples, as in \cite{loquercio18:article} or \cite{gandhi2017:article}, can be time-consuming, costly, and dangerous. We cannot afford potentially damaging collisions or causing injuries to nearby people. Our experiments show that the proposed methods outperform supervised baselines, and that small amounts of negative examples can improve traversability estimation in comparison to using only positive data, as in \cite{richter2017:inproceedings}. 

\begin{figure}[t]
  \centering
    \includegraphics[width=0.8\linewidth]{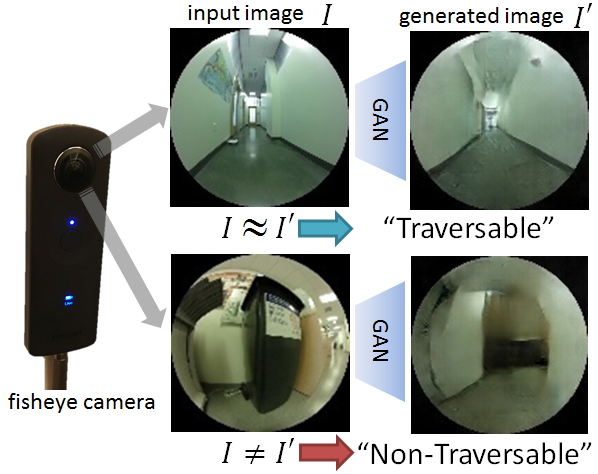}
	\caption{GONet takes as input an image of the environment, and estimates whether the space seen through the image is traversable.}
	\label{fig:header}
    \vspace{-1.5em}
\end{figure}

The first approach that we describe in this paper is GONet, a method that leverages powerful generative deep adversarial models \cite{goodfellow2014:inproceedings} to estimate traversability. The model takes as input a view of the environment from a fisheye camera on a robot, and predicts whether the area seen through the camera is safe to traverse (Fig. \ref{fig:header}). 
GONet is:
\begin{enumerate}[align=left,leftmargin=*,labelsep=0.4em,itemsep=0.4em]
\item \textit{cheap}, as it does not require expensive sensors, but just an off-the-shelf camera; 
\item \textit{fast}, because the structure of GONet is feed-forward, thus allowing for real-time operation on mobile platforms; 
\item \textit{robust}, because fisheye cameras efficiently capture every angle of the surrounding environment and our methods are capable of learning from a wide variety of data; and 
\item \textit{scalable}. Once trained on a given view, the model can run on other views from fisheye cameras positioned in a similar orientation with respect to the ground on the robot. For example, GONet can be used on a back-facing view after training on data from a forward-facing camera. 
\end{enumerate}

As part of our contributions, we propose two extensions to GONet. First, we show how it can be modified to enforce temporal consistency on traversability predictions.
To facilitate training on image sequences, we propose an automatic annotation procedure that increases the amount of labeled data without human supervision. Then, we extend our approach to work with stereo fish-eye cameras. While these extensions require a slight increase in computational power over our vanilla GONet model, they improve both the smoothness of our traversability predictions and their accuracy in challenging scenarios, such as when strong shadows significantly alter the appearance of the environment.

We conduct extensive experiments to validate the proposed semi-supervised methods, and contribute new datasets for data-driven traversability estimation in indoor environments. We also illustrate the applicability of our approach using real-time robot demonstrations. We show how GONet and its extensions can serve as an automatic visual emergency stop switch, enabling a robot to avoid collisions or falling down a flight of stairs. We also describe how our approach can complement 2D lidar range measurements during mobile robot navigation.

\section{PRELIMINARIES ON DEEP LEARNING}
\label{sec:preliminaries}

This section provides a brief introduction to Generative Adversarial Networks (GANs) \cite{goodfellow2014:inproceedings}, which are at the core of the proposed approaches for traversability estimation. Section \ref{sec:methods} then delves into the details of our methods.

GANs are a framework for estimating generative models through an adversarial
process. They simultaneously train two networks: a \textit{generator} ($Gen$) that captures the distribution of the training data and can produce examples from its manifold from a latent variable $z$; and a \textit{discriminator} ($Dis$) that tries to distinguish between samples generated by $Gen$ and actual samples from the training set.  GANs are considered unsupervised methods as they don't rely on labels, but use all the training data to learn both the generator and the discriminator. Training of $Gen$ is performed by means of a minimax two-player game, trying to maximize the probability of $Dis$ making a mistake. 

In this work, we use a particular class of GANs known as Deep Convolutional Generative Adversarial Networks (DCGANs) \cite{Radford15:article}. When the latent variable $z$ is sampled from some simple prior distribution, like a normal distribution, the generator of a DCGAN outputs an image $I' = Gen(z)$ that looks like the training data. Readers not familiar with GANs are encouraged to refer to \cite{goodfellow2016:arxiv} for an introductory tutorial.

\section{SEMI-SUPERVISED TRAVERSABILITY ESTIMATION}
\label{sec:methods}

This section introduces GONet, a new semi-supervised approach for traversability estimation from fisheye images. Then, we describe two extensions of the vanilla method: GONet+T, which enforces temporal consistency in the predictions; and GONet+TS, that uses stereo images while still considering temporal consistency. 

\subsection{GONet: Single-View Traversability Classification}
\label{sec:model_gonet}

GONet takes as input an image $I$ from a fisheye camera on a robot and outputs whether the environment seen through the image is traversable or not. The model is designed to learn from an uneven dataset with large numbers of positive examples, but just a few negative images of non-traversable areas. This capability is important because collecting negative data can be time-consuming, costly, and dangerous.

\begin{description}[align=left,leftmargin=0em,labelsep=0.5em,font=\textbf,itemsep=0.4em]

\begin{figure}[t]
  \centering
  \includegraphics[width=0.85\hsize]{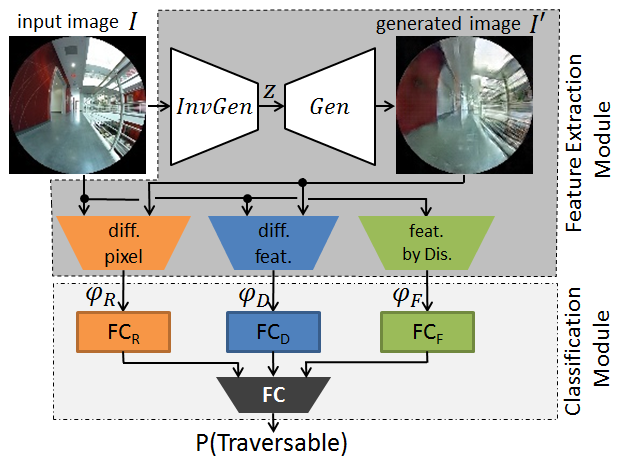}
  \caption{Overview of GONet. $I$ is the input image with dimensions $W \times H$, and $I'$ is the generated image (of the same size). $Gen$ corresponds to the generator of the DCGAN. $InvGen$ is an inverse generator to induce $I = I'$. 
  }
	\label{fig:GONet}
    \vspace{-1.5em}
\end{figure}

\item[Model Architecture.]
As shown in Fig. \ref{fig:GONet}, GONet is composed of two modules: a \textit{Feature Extraction Module} that is trained with automatically-labeled positive examples of traversable areas; and a \textit{Classification Module} that is trained with both positive and negative examples. Intuitively, GONet works by first generating an image $I'$  that is similar to the input $I$ and that looks as if it came from the manifold of positive examples, i.e., as if it belonged to the set of images of traversable areas.  The image $I'$ is generated with a DCGAN trained on positive examples only, as detailed in the next section. Then, GONet compares the input $I$ with the generated image $I'$ to decide whether the area seen through the input image is traversable. The main assumption of the model is that when the input indeed shows a traversable area, the generated image $I'$ would look very similar to it. But when the input depicts a non-traversable scenario, then the generated image would look different. 
%
%
More specifically, GONet estimates traversability from:
\begin{itemize}[align=left,leftmargin=*,labelsep=0.4em,itemsep=0.4em]
\item[--] $\phi_R = |I - I'|$, the residual difference  between the images,
 \item[--] $\phi_D = |f(I) - f(I')|$, the difference  between the discriminator features, and
 \item[--] $\phi_F = f(I)$, the discriminator features of the input image.
\end{itemize}
The features $f$ correspond to the last convolutional layer of the discriminator function of the DCGAN, which is trained  to distinguish between the true input $I$ and the generated $I'$.
\vspace{-.6em}

The features $\phi_R$ , $\phi_D$, and $\phi_F$ are processed by the Classification Module of GONet to output the traversability probability of the input image. First, these features are processed independently by fully connected layers that each output a scalar. These outputs are then concatenated into a $3 \times 1$ vector, and passed to another fully connected layer with a sigmoid activation function.
We optimize the final output using mean squared error and back-propagation. 

\item[Generating Images From the Positive Manifold.] GONet generates images from the positive manifold with a generator function $Gen$ from a DCGAN, as described in Sec. \ref{sec:preliminaries}. To ensure that the generated image $I'$ resembles the input $I$, we use another network to search for an appropriate variable $z$ that induces $I' = Gen(z)$ to be close to $I$ \cite{zhu16:inproceedings}. Note that this approach is faster than iteratively searching for the appropriate $z$ with back-propagation, as previously proposed for anomaly detection by Schlegl et al. \cite{schlegl17:inproceedings}.\vspace{0.4em} 


We create an auxiliary autoencoder \cite{bengio13:article} network as shown in the top part of Fig. \ref{fig:GONet} to find an appropriate variable $z$ for a given input image $I$. This autoencoder is composed of two modules: an inverse generator $InvGen$ that outputs $z$, and the generator $Gen$ from GONet's DCGAN. The inverse generator $InvGen$ is structured with the same layers as $Gen$, but in inverse order. The first four $InvGen$ convolutional layers have dimensions $64\times64\times64$, $32\times32\times128$, $16\times16\times256$ and $8\times8\times512$. The final layer is a fully connected layer with output size of $100$ (corresponding to the dimensionality of $z$). The auxiliary network for $InvGen$ is trained by  minimizing the loss $%
\mathcal{L}(z) = (1 - \lambda ) \mathcal{L}_R(z) + \lambda \mathcal{L}_D(z)
$, where $\mathcal{L}_R(z) = || I - Gen(z) ||$ is the residual loss for a given $z$, and $\mathcal{L}_D(z) =  ||f(I) - f(Gen(z))||$ is the discriminator loss. The features $f$  are the output of last convolutional layer of the DCGAN's discriminator. Note that, for training the $InvGen$ we freeze the $Gen$'s weights. The parameter $\lambda$ in $\mathcal{L}(z)$ trades-off between $\mathcal{L}_R(z)$ and $\mathcal{L}_D(z)$, and is chosen empirically with a the validation set.

\item[Data Collection \& Annotation.] To train GONet, we use data from a fisheye camera on a mobile robot. We drive this robot through safe traversable areas and automatically collect significant amounts of positive examples by inspecting the robot's velocity. If the robot moves continuously for $2.4$ s at a minimum velocity of 0.3 m/s, then we assume that the image collected in the middle of this time interval (at $1.2$s) depicts a traversable scenario. 
To gather a small set of negative examples, we carefully position the robot near obstacles and dangerous places. Section \ref{sec:datasets} details the datasets that we used for our evaluation.

\item[Training.] GONet is trained with back-propagation in 3 steps. First, we train a DCGAN with automatically annotated positive data. Through this GAN, we estimate the generator ($Gen$) and discriminator ($Dis$) of GONet. Second, we use the auxiliary autoencoder network shown at the top part of Fig. \ref{fig:GONet} to train the $InvGen$ using positive examples. Third, we train the final fully connected layer of GONet with a small set of positive and negative examples. We use early stopping in this last step to prevent over-fitting.

\end{description}

\begin{figure}[t]
  \centering
    \includegraphics[width=0.95\hsize]{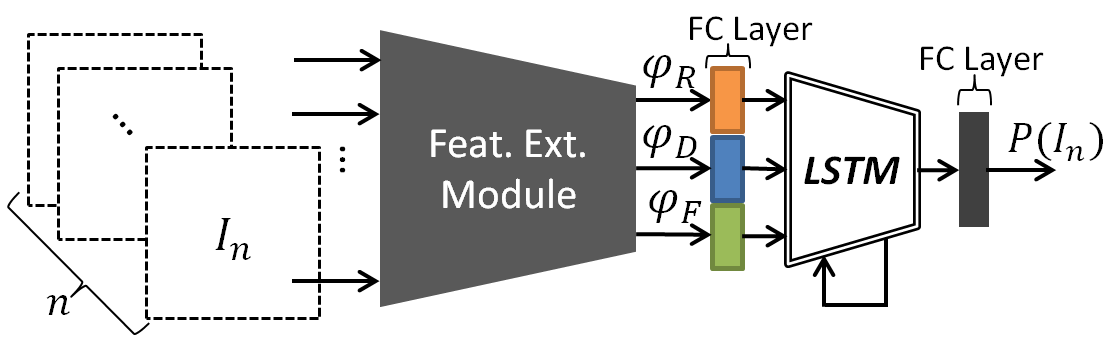}
	\caption{General network structure for GONet+T and GONet+TS. The Feature Extraction Module is the same as in Fig. \ref{fig:GONet}, except that now the input is a tensor with dimensions $W \times H \times n$. GONet+T receives as input a color image, just like GONet. This corresponds to $n=3$ for the RGB channels. GONet+TS processes stereo color images, corresponding to $n=6$.}
    \label{fig:nchannel-gonet}
        \vspace{-1.5em}
\end{figure}

%
%

\subsection{GONet+T: Enforcing Temporal Consistency}
\label{sec:gonet+t}
GONet assumes that every input image is independent and identically distributed; but this is not true for image sequences captured from a robot \cite{darrell15:inproceedings}. Current observations are dependent on the past, and this past can often help understand whether the present state is traversable or not.

We propose GONet+T to account for the temporal nature of data gathered from a robot. GONet+T  extends GONet by using a Long Short-Term Memory (LSTM) unit \cite{gers99:article} to reduce the variance on traversability predictions over time. LSTMs are popular Recurrent Neural Network models capable of learning long-term dependencies. These recurrent models have shown great promise for sequential tasks \cite{lecun2015deep}.


\begin{description}[align=left,leftmargin=0em,labelsep=0.5em,font=\textbf,itemsep=0.4em]
\item[Model Architecture.] 
GONet+T first uses the Feature Extraction Module of GONet to compute $\phi_R$, $\phi_D$, and $\phi_F$, as depicted in Fig. \ref{fig:nchannel-gonet} for $n=3$. The features are then  transformed in GONet+T by three fully connected layers to reduce their dimensionality to $10$-dimensional vectors each. These vectors are concatenated and, consequently, processed by a recurrent LSTM unit. The LSTM learns temporal dependencies in the data, and its output is finally passed to a fully connected layer with a sigmoid activation function to predict the traversability probability.
%

We optimize GONet+T with a multi-objective loss function: 
$\mathcal{L} =  \lambda \sum_{i=0}^{T}{||y_i - {y'}_i||} + (1 - \lambda) \sum_{i=0}^{T-1}||{y'}_{i+1} - {y'}_i||$, 
%
%
where $y_i$ corresponds to the ground truth traversability probability at time $i$, ${y'}_i$ and ${y'}_{i+1}$ are the predicted probabilities at times $i$ and $i+1$, respectively, and $T$ is the length of the training image sequences -- here we assume equal lengths for simplicity. The $\lambda$ parameter trades-off between the prediction error $||y_i - {y'}_i||$ and the smoothness term $||{y'}_{i+1} - {y'}_i||$. We choose $\lambda=0.5$ using the validation set.


\item[Data Collection \& Annotation.] The use of an LSTM unit in GONet+T requires fully annotated sequences at training time. But annotating vast amounts of negative examples in these sequences is costly and very time consuming. Thus, we devised the following semi-automatic annotation procedure for fisheye image sequences collected from a robot:
\begin{enumerate}[align=left,labelsep=0.5em,itemsep=0.4em]
\item We collect data by driving the robot through safe places. The robot is controlled to carefully approach and stop when its near obstacles and dangerous spaces.
\item We annotate vast amounts of positive data based on the robot's velocity (as mentioned before), and manually label a small amount of randomly-chosen negative samples.
\item We train GONet (as in Sec. \ref{sec:model_gonet}) with the labeled data that we have so far. We use all positive data for GONet's Feature Extraction Module, and a small set of positive and negative examples for the Classification Module.
\item We run GONet on the remaining images that were collected while the robot was static or moved at low speeds, and label these images with GONet's output probability.
\end{enumerate}

To prove that this annotation procedure would not negatively affect training (as GONet might naturally make mistakes), we conducted an empirical evaluation on the predictions of GONet before and after the re-evaluation. The mean difference between the two sets of predictions for the entire unlabeled dataset is calculated as $\epsilon = 1/|U| \times \sum_{i\in U}|P(i)-P'(i)|$. Where $P(i)$ is the output probability of GONet, $P'(i)$ is the output probability of GONet after re-training on the re-annotated data, and $U$ is the set of all unlabeled data. 
After the re-annotation, the accuracy of our network on the positive training set increased because they are correctly labeled through the robot's moving experience. We show that $\epsilon = $ 0.04, showing that the predictions of GONet before and after training on the re-annotation dataset does not change.

\item[Training.] We train GONet+T with the same procedure that we used to train GONet. Except that in this case we optimize the final output of the model using labeled image sequences. 

\end{description}

\subsection{GONet+TS: Estimating Traversability With Stereo Views}

While GONet and GONet+T worked well in different situations, their performance was sometimes affected by the appearance of the environment, e.g.,  strong shadows occasionally induced false negative predictions. As a result, we decided to extend GONet+T to reason about stereo fisheye images. Our hypothesis was that having two views of the environment would increase the robustness of our traversability estimation approach in challenging scenarios. 

GONet+TS has the same architecture as GONet+T, except for the fact that it receives as input a 6 channel tensor, as illustrated in Fig. \ref{fig:nchannel-gonet} for $n=6$. The first three channels of the input tensor correspond to the RGB channels of the left image in the stereo pair; the last three channels correspond to the right image. We collect data and annotate it as in the previous section. We also train GONet+TS with the same loss and in the same fashion as GONet+T. 

\section{EVALUATION}
\label{sec:evaluation}

This section describes the experiments to evaluate the proposed methods in indoor environments. Results were computed using an NVidia GeForce GTX TITAN X GPU.

\subsection{Datasets}
\label{sec:datasets}

We collected two datasets to evaluate the proposed traversability estimation methods.\footnote{We will open-source our datasets upon acceptance.} The first dataset, ``Go Stanford 1'' (GS1), was collected as described in Sec. \ref{sec:model_gonet}. We teleoperated a TurtleBot2 robot\footnote{\url{https://www.turtlebot.com/turtlebot2/}} with an onboard Ricoh THETA S fisheye camera\footnote{Although the THETA S has two fisheye cameras (one in the front and one in the back), we only use the front camera to
capture the environment. The horizontal and vertical fields of view of this camera are both $180^\circ$.} in 15 different buildings at Stanford University (totaling 7.2 hours of sensor data).  For each data collection session, we recorded the view from the robot's fisheye camera in HD resolution and at 3Hz. We also collected the teleoperation commands, and robot's velocity. The velocity was used to automatically annotate 35783 positive examples, out of a total of 78711 useful images. In our experiments, we split the data to $9$, $3$, and $3$ buildings for training, validation, and testing, respectively. 


The second dataset, ``Go Stanford 2'' (GS2), was recorded with a TurtleBot2 with two fisheye THETA S cameras. The cameras were placed in front of the robot, with a baseline distance of $0.115$m. We collected the same data as for GS1 in GS2, but this time we operated the robot in 27 campus buildings. 
GS2 is composed of a total of 16.74 hours of video ($177297$ useful images). We split the data to $19$ buildings for training, $4$ for validation, and $4$ for testing.

We annotated $2400$ negative examples in GS1 and GS2 ($400$ images for training, $400$ for validation, and $400$ for testing in each dataset). We then complemented this data with the same amount of automatically annotated positive images. We also augment GS1 by flipping images horizontally, and GS2 by swapping and flipping the stereo images. For all the experiments we resize the images to 3$\times$128$\times$128.


\subsection{Data efficiency}
We first investigate the performance of GONet as the number of annotated positive and negative examples grows. We first train the Feature Extraction Module of GONet with the positive data from GS1. Then, we train GONet's Classification Module with $100$, $200$, $400$, $800$, $1600$ and $2400$ training examples from GS1 and GS2. We evaluate the models on the unseen labeled images from GS1's testing set. 

{\small
\begin{table}[b]
  \centering
  \caption{Accuracy vs. Num. of positive \& negative training examples}
  \label{tab:data}
  \resizebox{\columnwidth}{!}{%
   \begin{tabular}{c||c|c|c|c|c|c}\wcline{1-7}
       Training data & 100 & 200 & 400 & 800 & 1600 & 2400  \\ \wcline{1-7}
       Accuracy [$\%$] & 89.12 & 91.5 & 92.12 & \textbf{94.25} & \textbf{94.875} & \textbf{94.750} \\ \wcline{1-7}
	\end{tabular}
    }
\end{table}}

As can be seen in Table \ref{tab:data}, GONet  is  data efficient. It quickly reaches $94.25\%$ accuracy using only 800 training examples ($400$ positive and $400$ negative images). For this reason, we use $800$ images for training supervised models in the experiments in the following sections.


\subsection{GONet Evaluation on GS1}
\label{ssec:gs1-results}

Using GS1, we performed an ablation study for GONet and evaluated it against  fully supervised and unsupervised baselines for traversability estimation. The unsupervised models or components of the methods considered in this evaluation were all trained with the same positive data from GS1. The supervised models were trained on the same $400$ positive and $400$ negative training examples, and also used the same validation set. Classification thresholds $\tau$ for unsupervised methods were chosen using the validation set; the thresholds for supervised methods were optimized with the training data.
We considered the following models: 
\begin{description}[align=left,leftmargin=0em,labelsep=0.5em,font=\raisebox{0.25ex}{\tiny$\bullet$}\normalsize~\textnormal,itemsep=0.4em]

\item[Autoencoder:] Following \cite{richter2017:inproceedings}, we use an autoencoder as an unsupervised method to classify traversability. The autoencoder has the same structure as the auxiliary network used to train $InvGen$ (see Sec. \ref{sec:model_gonet}). However, we train this autoencoder from scratch to minimize the $\ell^2$ difference between the input and the generated images. Traversability is decided based on $\sum{\|I - I'\|^2} < \tau$.

\item[Resnet:] As a baseline, we try predicting traversability with state-of-the-art ResNet 50 and ResNet 152 features for image recognition \cite{he2016deep}. The ResNet models are pre-trained on ImageNet, and their features for a given input image are evaluated by a fully connected layer to predict traversability. To avoid over-fitting, we only train the final layer.

\item[Feature Extraction Module:] We compare the proposed approach to estimate $InvGen$ in GONet's Feature Extraction Module with the iterative back-propagation method of \cite{schlegl17:inproceedings}. We  threshold image reconstruction errors for these models to classify the input images, as we did for the autoencoder.  


\item[GONet:] We break-down the contributions of GONet's components. Based on the features output by its Feature Extraction Module, we consider classifying traversability using the $\phi_R$ feature (R Model), the $\phi_D$ feature (D Model), the $\phi_f$ feature (F Model), and their combinations. Classification thresholds are chosen with a final fully connected layer appended to individual features. For their combinations, we append two fully connected layers as in Fig. \ref{fig:GONet}.
\end{description}

Table \ref{tab:analysis1} shows our results on GS1. The top section of the table shows the performance, speed, and memory requirements of the models trained with positive examples only (P); the bottom section shows results with both positive and negative examples (P+N). In general, using a small amount of negative examples helped improve accuracy.

By comparing rows 1-3 of Table \ref{tab:analysis1} (excluding the header), we can observe that GONet's Feature Extraction model outperforms the baseline autoencoder \cite{richter2017:inproceedings}. While these two methods work in a similar spirit, our DCGAN with the inverse generator results in highest accuracy across the unsupervised methods that were trained with positive data only. In terms of accuracy and speed, the results also show the effectiveness  of using the inverse generator \cite{zhu16:inproceedings} in GONet in contrast to using iterative back-propagation \cite{schlegl17:inproceedings}.

The results in the bottom section of Table \ref{tab:analysis1} suggest that each feature output by GONet's Feature Extraction Module and the components of its Classification Module contribute to the accuracy of the proposed approach. In addition, the full GONet architecture (R+D+F GONet Model in the last row) outperforms all other models evaluated on GS1. In contrast to ResNet, not only does GONet lead to higher accuracy, but it also reduces memory requirements and computation time significantly. The reason is that while ResNet shines at image recognition tasks due to its deep architecture, the network's size also requires more computational resources. Since mobile robots must often depend on less powerful onboard computing hardware, lower memory usage and faster computation time are desirable qualities for our application. 

{\small
\begin{table}[t]
  \centering
  \caption{Results on GS1's test set. See Sec. \ref{ssec:gs1-results} for more details.}
  \label{tab:analysis1}
  \resizebox{\columnwidth}{!}{%
   \begin{tabular}{c|c|c|c|c|c}\wcline{1-6}
       & Trained on & Model & Accuracy [$\%$] & Hz & Memory [MB]  \\ \wcline{1-6}
       baseline  & P & Autoencoder \cite{richter2017:inproceedings} & 64.25 & \textbf{205} & 557 \\ \wcline{1-6}
       Feat. Ext. Module  & P &  back-prop \cite{schlegl17:inproceedings} & 60.00 & 0.125 & \textbf{323} \\ \cline{2-6}
        & P &  $InvGen$ & \textbf{72.50} & 93.07 & 354 \\ \wcline{1-6} \hline \hline \wcline{1-6}
       baseline  & P+N & ResNet 50 \cite{he2016deep} & 91.63 & 34.46 & 705 \\ \cline{2-6}
       & P+N & ResNet 152 \cite{he2016deep} & 92.25 & 12.21 & 1357  \\ \wcline{1-6}
       & P+N & R & 85.38 & 175.17 & 338  \\ \cline{2-6}
       & P+N & D & 91.63 & 103.17 & 356 \\ \cline{2-6}
       & P+N & F & 92.25 & \bf{329.37} & \textbf{326}  \\ \cline{2-6}
       GONet & P+N & R+D & 91.63 & 94.11 & 358  \\ \cline{2-6}
       & P+N & D+F & 93.00 & 96.41 & 357  \\ \cline{2-6}
       & P+N & R+F & 93.13 & 119.99 & 348  \\ \cline{2-6}
       & P+N & \bf{R+D+F} & \bf{94.25} & 89.69 & 359  \\ \wcline{1-6}
	\end{tabular}
    }
\vspace{-1.5em}

\end{table}}

\subsection{Evaluation of GONet and its extentions on GS2}

\label{ssec:gs2-results}
\begin{table*}[h]
  \centering
  \caption{Results on GS2's test set. See Sec. \ref{ssec:gs2-results} for more details.}
  \label{tab:analysis}
  \begin{tabular}{l|l|l|c|c|c|c|c|c} \wcline{1-9}
       Model & Sensor & Training Data & Neg. accu.[$\%$] & Pos. accu.[$\%$] & Accu. [$\%$] & Prec. [$\%$] & Hz & Mem. [MB] \\ \hline \cline{1-9}
       (a): 4ch GONet & Kinect & Hand Labeled & 89.60 & 87.20 & 88.40 & 89.34 & 75.75 & 416\\ \cline{1-9}
       (b): disparity map & two THETA S & Hand Labeled &  66.20 & 75.20  & 70.70 & 68.99 & - & - \\  \cline{1-9}
       (c): siamese net & two THETA S & Hand Labeled & 88.40 & 94.20 & 91.30 & 88.98 & 8.63 & 3744 \\ \hline \cline{1-9}        
       (d): GONet & one THETA S & Hand Labeled & 90.60 & 94.65 & 92.55 & 90.63 & \textbf{110.80} & \textbf{339} \\ \cline{1-9}
       (e): (d) + re-annotation & one THETA S & Re-annotation & 90.6 & 96.20 & 93.40 & 91.10 & $\uparrow$ & $\uparrow$ \\ \cline{1-9}
       (f): GONet+T & one THETA S & Re-annotation & 91.7 & 97.20 & 94.45 & 92.05 & 107.41 & 347 \\ \hline \cline{1-9}
       (g): GONet+S & two THETA S & Hand Labeled & 93.80 & 96.00 & 94.90 & 93.93 & \textbf{109.24} & \textbf{413} \\ \cline{1-9}       
       (h): (g) + re-annotation & two THETA S & Re-annotation & 95.80 & 96.20 & 96.00  & 95.82 & $\uparrow$ & $\uparrow$ \\ \cline{1-9}       
       (i): GONet+TS & two THETA S & Re-annotation & \textbf{96.20} & \textbf{97.60} & \textbf{96.90}  & \textbf{96.26} & 86.761 & 429 \\ \hline \cline{1-9}
	\end{tabular}
        \vspace{-0.5em}

\end{table*}

In this section, we evaluate GONet and its two extensions, GONet+T, and GONet+TS, against baselines in the GS2 dataset. We take advantage of the two cameras used to collect images in GS2 to perform comparisons across systems based on accuracy, speed, and memory requirements. All the models considered in this experiment are evaluated on GS2's testing set. We choose classification thresholds 0.5 and model parameters based on the validation set. This evaluation considered the following models:
%
%
\begin{description}[align=left,leftmargin=0em,labelsep=0.5em,font=\raisebox{0.25ex}{\tiny$\bullet$}\normalsize~\textnormal,itemsep=0.4em]

\item[4ch GONet with Kinect:] We train a GONet with a 4 channel input tensor: the first 3 channels correspond to the Kinect's RGB image and the last channel corresponds to the Kinect's depth map.


\item[Disparity Map:] We compute disparity maps \cite{hirschmuller2008stereo} using the left and right image pairs in GS2. The depth maps is reshaped into a vector, and passed to two fully connected layers to predict traversability.

\item[Simese Net:] In the spirit of \cite{kendall2017end}, we explored using a siamese network with ResNet 152 features. We concatenate the ResNet features from the image pairs and pass them two three fully connected layers to classify traversability. In this baseline, we only train the fully connected layers.


\item[GONet Models:] We evaluate all the proposed methods (GONet, GONet+T, and GONet+TS) for traversability estimation after training on hand labeled annotations only ($400$ positive and $400$ negative examples) and after re-annotated data as explained in Sec. \ref{sec:gonet+t}.


\end{description}

Table \ref{tab:analysis} shows the results on GS2. The annotation process proposed in Sec. \ref{sec:gonet+t} increased the accuracy. By comparing the results in rows (f) and (i) of Table \ref{tab:analysis} with the other rows, we can see that the use of LSTM units in GONet also increased performance. We attribute this success to the fact that GONet+T and GONet+TS are able to reason about the inter-dependency of the input images.


Our GONet model and its extensions outperformed the supervised baselines that used depth maps (rows (a) and (b)) and pre-trained features (row (c)). In particular, GONet+TS had the best performance of all the models on GS2. This provided support to our hypothesis that reasoning about stereo fisheye images can be help estimate traversability.


\subsection{Qualitative Analysis}
\begin{figure}[t]
\centering
    \begin{tabular}{cc}
      \includegraphics[width=0.27\hsize]{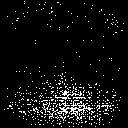} & \includegraphics[width=0.27\hsize]{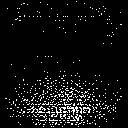} 
    \end{tabular}
\caption{Mean saliency map of GONet+TS for the left and right images.}
\label{f:saliency}
\vspace{-1.5em}
\end{figure}

\begin{figure*}[t]
  \centering
    \begin{tabular}{cc}
      \includegraphics[width=0.48\hsize]{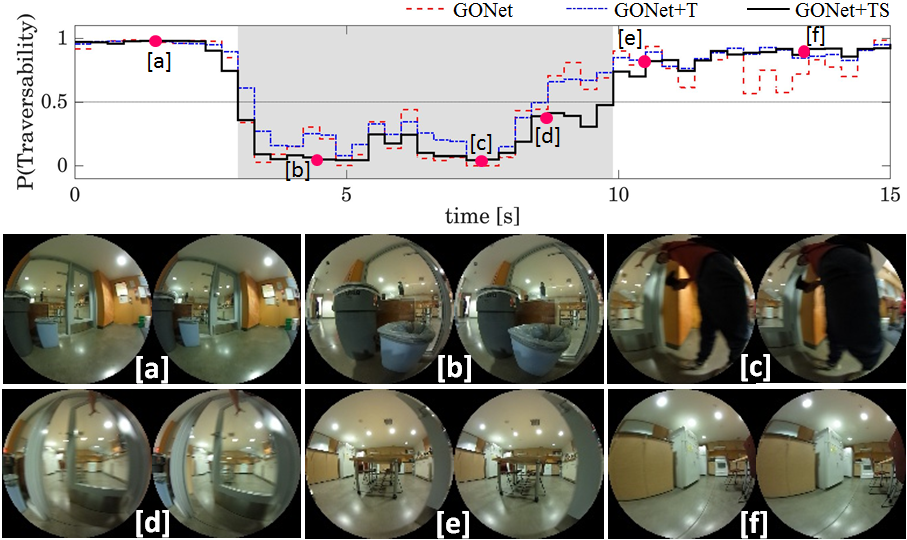} & \includegraphics[width=0.48\hsize]{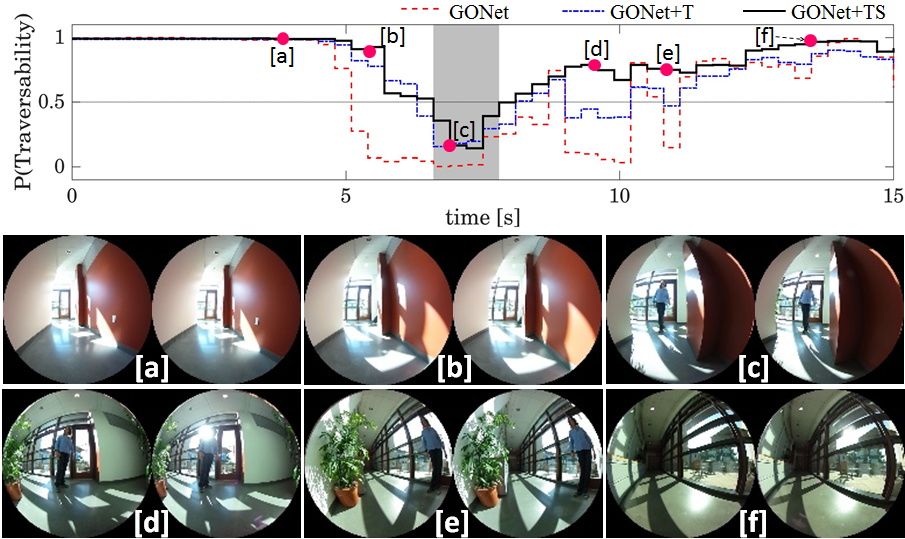} \\
      $\left( \mbox{i} \right)$ passing through the door & $\left( \mbox{ii} \right)$ sun light and shade \\
    \end{tabular}
\caption{Output of GONet and its extensions for 2 challenging scenarios from GS2 test set and the corresponding stereo images of different time points.}
\label{f:res}
    \vspace{-0.7em}
\end{figure*}

\begin{description}[align=left,leftmargin=0em,labelsep=0.5em,font=\textbf,itemsep=0.4em]
\item[Generated Images.] Fig. \ref{f:generated} shows several example input images $I$ and corresponding generated images $I'$ by GONet's Feature Extraction Module. Because this module was trained only on positive examples, the obstacles in the generated images are often blurred out. Likewise, closed pathways in the input images tend to be converted to open spaces. These changes are what enables the proposed approaches to identify unsafe and non-traversable areas.

\item[Saliency Maps.] We use saliency maps \cite{Simonyan17:article} to visualize the behavior of GONet+TS after training on GS2. As shown in Fig. \ref{f:saliency}, the bottom part of the input images were the most salient and, thus, the most relevant for the classification task. These areas correspond to space in front of the robot. 


\item[Traversability Predictions.] Fig. \ref{f:res} visualizes GONet, GONet+T and GONet+TS predictions in two challenging scenarios. The gray areas in the plots correspond to time periods in which the environment right in front of the robot was unsafe to traverse. In particular, Fig. \ref{f:res}[i] depicts a situation in which a robot enters a room with a glass door. At the beginning, the door is closed (images [a,b,c]) and, then, a person opens the door and holds it for the robot to pass.  In general, all the methods tended to perform well in this situation. GONet+T and GONet+TS led to smoother predictions in comparison to GONet thanks to their LSTM.

Fig. \ref{f:res}[ii] shows a situation in which the robot moves through areas with strong shadows. At time [b], GONet mistakenly predicts that the area in front of the robot is unsafe to traverse. At time [d], both GONet and GONet+T output false negatives; only GONet+TS can correctly predict traversability. In our experiments, enforcing temporal consistency in the traversability predictions and using stereo vision often helped  deal with such challenging scenarios.

\begin{figure}[t]
   \centering
   \includegraphics[width=0.9\hsize]{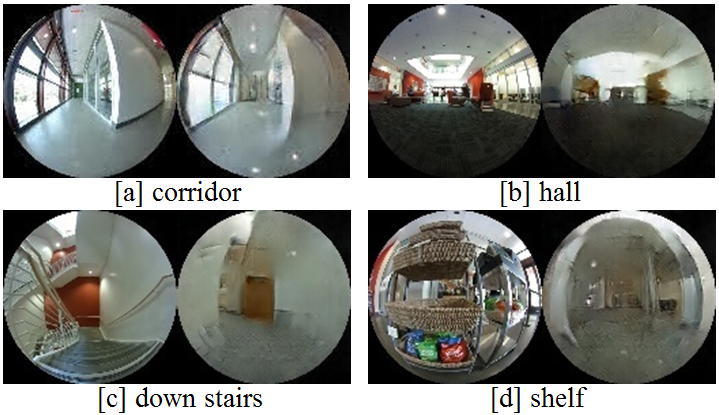}
 	\caption{In each image, input images $I$ (left) and generated images $I'$ (right) by GONet's Feature Extraction Module. The top row corresponds to traversable areas; the bottom depicts non-traversable areas.}
 	\label{f:generated}
\vspace{-0.8em}
 \end{figure}

\end{description}


\section{APPLICATIONS}
We demonstrate how the proposed approaches for traversability estimation can be used as an autonomous visual emergency stop mechanisms for mobile robots. We also argue that these approaches can increase the robustness of the standard (2D) lidar-based exploration and mapping methods. Finally, we demonstrate the versatility of GONet by showing how it can be used as a navigation aid for visually impaired people. Due to space constraints, we omit results with GONet+TS in this paper (even though we implemented  systems equivalent to those discussed in the following sections with stereo cameras). We also mainly discuss GONet+T as it was more accurate than GONet in our quantitative evaluation (Sec. \ref{sec:evaluation}). Additional supplementary material for these demonstrations can be found in the accompanying video submission and online.\footnote{\url{http://cvgl.stanford.edu/gonet}}

For all the demonstrations described in this Section, we ran GONet and GONet+T on a laptop with an Intel Core i7-6700HQ processor, 32GB of RAM, and an NVidia GeForce 980M GPU. The laptop was connected to a Ricoh THETA S fisheye camera positioned at the front of a TurtleBot2 robot (Sec. \ref{ssec:estop} \& \ref{ssec:app-nav}) or manually held by a user (Sec. \ref{ssec:blind}). GONet and GONet+T could run on the laptop at the maximum frame rate of the camera (15Hz), but we limited how fast they processed images to match the frame rate used at training time (3Hz). 


\subsection{Visual Emergency Stop (E-Stop) Switch}
\label{ssec:estop}

GONet+T ran on a teleoperated TurtleBot2 to signal unsafe areas in front of the robot. When these areas were detected, GONet+T overrode teleoperation commands to force the robot to stop. Our traversability estimation approach was able to prevent the robot from falling down stairs, and it stopped the robot from colliding with glass walls, fences, doors, and people in previously unseen environments. 
Moreover, we tested the ability of GONet+T to recognize new obstacles including a piece of aluminum, a jacket, and a binder that were suddenly thrown in front of the robot.

While GONet+T was able to prevent the robot from falling and colliding in many different situations, we observed an interesting failure case with a tangled wired on the ground. When the robot approached the wire, GONet+T was able to stop the TurtleBot, but it did so late. Part of the wire was already underneath the robot. A few other failure cases happened with very small objects, like a computer mouse lying on the ground that had a color similar to the carpet. GONet+T missed these few objects, and the TurtleBot pushed them. 

\subsection{Mobile Robot Exploration \& Mapping}
\label{ssec:app-nav}
\begin{figure}[t]
   \centering
   \includegraphics[width=1.0\hsize]{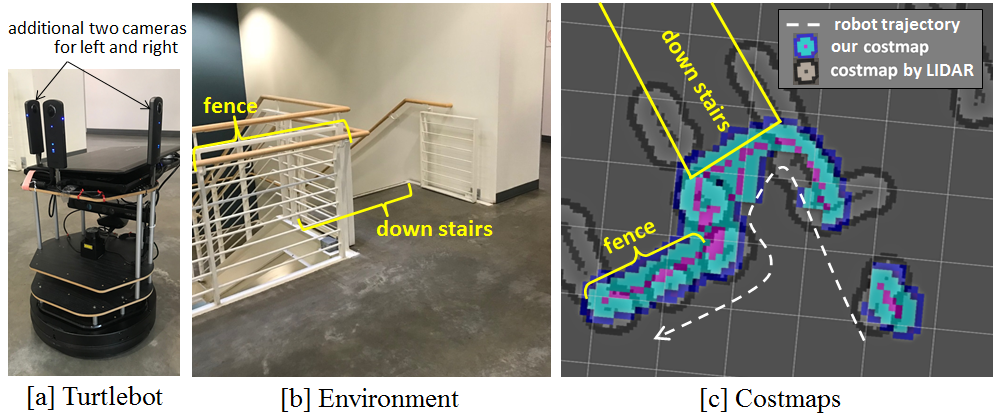}
 	\caption{Turtlebot with 3 fisheye cameras (a), environment where we tested GONet+T (b), and overlaid costmaps created with GONet+T and lidar (c).}
 	\label{f:costmap}
 \vspace{-1.5em}
 \end{figure}

We added a 2D lidar to the TurtleBot2 that we used for our applications to enable the robot to navigate and explore the environment with standard probabilistic mapping methods \cite{thrun2005:book}. 
We also added two more fisheye cameras on the robot to demonstrate the scalability of GONet at handling different views of the environment (see Fig. \ref{f:costmap}). Based on the traversability predictions from GONet, we automatically updated the robot's costmap as it navigated inside a building in the test set of our GS1 dataset.  
GONet identified blocked and unsafe areas that the lidar missed, such as parts of a fence and stairs (Fig. \ref{f:costmap}). Because recognizing these critical areas is essential for autonomous robot operation, we believe that the proposed traversability estimation methods can increase the robustness of current navigation approaches. Investigating effective mechanisms to combine GONet with other sensing modalities is an interesting future research direction.

\subsection{Assistive Traversability for the Visually Impaired}
\label{ssec:blind}

We  implemented a navigational aid system that produces an audible warning when an obstacle or drop-off is detected by GONet+T, and placed the laptop with our system in a backpack. We then asked a person to carry this backpack and to hold the fisheye camera that was connected to the laptop so that it could gather images of the environment in front of the person. 
As this person walked blindfolded through unseen indoor and outdoor environments, GONet+T recognized unsafe areas, and our system correctly issued audible warnings. Note that we never trained GONet+T in outdoor environments, nor with images gathered by a pedestrian. Nonetheless, our traversability estimation approach showed great generalization potential under new conditions.

As cameras and computational devices continue to shrink in size, such a warning system could be mounted in a small enclosure either on a belt or even integrated into a pair of glasses. We are excited about the porential of GONet to advance application areas beyond traditional robotics. 



\section{RELATED WORK}
\label{sec:related_work}

The traversability estimation approaches presented in this work were motivated by the success of deep learning on many visual tasks \cite{lecun2015deep}. By leveraging deep generative adversarial models \cite{goodfellow2014:inproceedings, Radford15:article}, our methods were able to process raw images to distinguish traversable and non-traversable areas. One advantage of these methods is that they do not rely on hand-tuned image features, as some prior efforts \cite{ulrich00:inproceedings, batavia01:inproceedings, lenser2003:inproceedings}. Rather, the proposed methods learn from data the relevant information for the traversability estimation task. 

Several prior efforts suggested using only positive data for traversability estimation \cite{ulrich00:inproceedings, suger15:inproceedings, richter2017:inproceedings}. Inspired by this line of work, we collected vast amounts of positive examples by driving a robot through safe places, and used this data to train our models. However, in contrast to these other efforts, we also collected a small number of negative images. These negative examples increased the accuracy of GONet in comparison to using positive examples only,  as in \cite{richter2017:inproceedings}.

At a high level, the proposed approaches can be considered anomaly detection methods \cite{chandola09:article}. GONet and its extensions expect to get as input an image that depicts a traversable area; negative examples are outliers -- they correspond to non-conforming patterns.
In contrast to prior work in deep learning on anomaly detection \cite{schlegl17:inproceedings}, our methods use a visual manipulation technique \cite{zhu16:inproceedings} to generate images from the positive manifold. The generated images not only look like traversable areas, but also resemble the input query. This makes our approach fast and practical for mobile robots. To the best of our knowledge, this is the first time that this image manipulation technique is used for traversability estimation. 

It is worth noting that prior work involving learned motion control policies for robots have started to reason about the traversability of environments along with a robot's next actions \cite{dey2016:inbook, giusti2016machine, richter2017:inproceedings, loquercio18:article}. While combining traversability estimation and policy learning is out of the scope of this paper, our experimental results suggest that GONet and its extensions can potentially facilitate learning motion behaviors for robots. For example, GONet could be combined with a convolutional network to predict steering angles, as in \cite{loquercio18:article}.

\section{CONCLUSION \& FUTURE WORK}
\label{sec:conclusion}
We presented semi-supervised methods for traversability estimation. At the core of these methods are powerful deep generative models that distinguish between traversable and non-traversable areas. The models are trained with vast amounts of positive images depicting safe areas that a robot can navigate through, but just a few negative examples. The ability to learn from such uneven data makes our methods more robust than supervised and fully unsupervised baselines while keeping them practical. GONet and its extensions are cheap and fast. They take as input, images from fisheye cameras and process them in a feed-forward fashion, thus allowing for real-time operation on mobile hardware. We demonstrated how the proposed methods can save a robot from dangerous situations, e.g., falling down stairs and colliding with glass. Moreover, this same capability can be used to complement 2D lidar range measurements during robot navigation and exploration.  Finally, we demonstrated how the trained GONet can be removed from the robot and carried by a visually impaired individual to issue warnings when the user is heading toward obstacles or drop-offs.

Our experiments focused on traversability estimation in indoor environments. Future work should explore the accuracy of the proposed methods on outdoor settings. Another interesting avenue of future research is end-to-end joint traversability and policy learning. GONet and its extensions can potentially facilitate learning robust motion behaviors. 



\bibliographystyle{IEEEtran}
\bibliography{IEEEabrv,references}

\end{document}